\documentclass{article}

\usepackage{PRIMEarxiv}

\usepackage[utf8]{inputenc} % allow utf-8 input
\usepackage[T1]{fontenc}    % use 8-bit T1 fonts
\usepackage{hyperref}       % hyperlinks
\usepackage{url}            % simple URL typesetting
\usepackage{booktabs}       % professional-quality tables
\usepackage{amsfonts}       % blackboard math symbols
\usepackage{nicefrac}       % compact symbols for 1/2, etc.
\usepackage{microtype}      % microtypography
\usepackage{lipsum}
\usepackage{fancyhdr}       % header
\usepackage{graphicx}       % graphics
\graphicspath{{media/}}     % organize your images and other figures under media/ folder
\usepackage{amsmath}
\title{End-to-end Semantic Object Detection with Cross-Modal Alignment
}

\author{
  Silvan Ferreira, Allan Martins, Ivanovitch Silva \\
  Federal University of Rio Grande do Norte, Natal, Brazil \\
  \texttt{silvan.junior.051@ufrn.edu.br, allan@dee.ufrn.br, ivanovitch.silva@ufrn.br} \\
}

\begin{document}
\maketitle

\begin{abstract}
Traditional semantic image search methods aim to retrieve images that match the meaning of the text query. However, these methods typically search for objects on the whole image, without considering the localization of objects within the image. This paper presents an extension of existing object detection models for semantic image search that considers the semantic alignment between object proposals and text queries, with a focus on searching for objects within images. The proposed model uses a single feature extractor, a pre-trained Convolutional Neural Network, and a transformer encoder to encode the text query. Proposal-text alignment is performed using contrastive learning, producing a score for each proposal that reflects its semantic alignment with the text query. The Region Proposal Network (RPN) is used to generate object proposals, and the end-to-end training process allows for an efficient and effective solution for semantic image search. The proposed model was trained end-to-end, providing a promising solution for semantic image search that retrieves images that match the meaning of the text query and generates semantically relevant object proposals.
\end{abstract}

% keywords can be removed

\section{Introduction}
The ability to detect and categorize objects is a crucial component of human perception and cognitive function. This process is often guided by selective attention, allowing us to efficiently search for and locate objects of interest in our surroundings while ignoring other irrelevant stimuli \cite{desimone1995neural}. In computer vision, the task of object detection aims to replicate this ability by automatically identifying and locating objects in images and videos. However, traditional object detection methods often lack the semantic understanding and attentional mechanism present in human object search, leading to limitations in their performance.

Object detection establishes itself with great importance in the field of computer vision, which focuses on identifying and locating objects within an image or video. This problem has been actively researched for decades and has recently gained a surge of attention due to its various real-world applications. Some of the most prominent applications include self-driving cars, surveillance systems, and human-computer interaction. These applications require the accurate detection of objects in real-time, making it an essential tool for many industries \cite{zou2023object}.

The task of object detection can be divided into two subtasks, object proposals generation and classification. The former involves generating candidate regions or proposals within an image or video frame that may contain objects, while the latter is the task of classifying each proposal as foreground or background. This process requires the integration of both subtasks to generate accurate and efficient results.

Despite the advances made in object detection, it remains a challenging task due to the large variability in object appearance and scale, as well as the presence of clutter and occlusions. Additionally, existing object detection models are not designed to take into account the semantic relationships between objects and the text queries used to search for them \cite{zou2023object}.

Semantic image search also plays a significant role in computer vision by allowing users to search for images based on their semantic content. Unlike traditional search methods that rely solely on keyword matching, semantic image search leverages the understanding of visual and textual information to return results that are semantically relevant to a given query. This is an important development in the field of computer vision, as it enables a more natural and intuitive way of searching for visual information, and makes it possible to retrieve images based on the meaning they convey, rather than just their surface features. The common approach for this task is training a cross-modal model to align visual and textual information, as in \cite{qi2020imagebert,radford2021learning}. Nevertheless, traditional methods of semantic image search matches a text input with the whole image, without considering the location of the objects within the image. Which narrows the ability of these approaches to handle real-world scenarios where a user may need to search for objects in a specific image.

In this paper, we present a novel approach to object detection that addresses the challenges in semantic image search. Our proposed model aims to improve the semantic alignment between the text and object proposals by considering the semantic relationship between the two modalities. The proposed model consists of a single feature extractor, a pre-trained Convolutional Neural Network \cite{li2021survey}, and a Transformer encoder \cite{vaswani2017attention} to encode the text query. Proposal-text alignment is performed using contrastive learning \cite{jaiswal2020survey}, producing a score for each proposal that reflects its semantic alignment with the text query. The Region Proposal Network (RPN) \cite{ren2015faster} is used to generate object proposals, and the end-to-end training process allows for an efficient and effective solution for semantic image search.

The contribution of this paper is two-fold. First, we introduce a novel approach to object detection that considers the semantic alignment between object proposals and text queries. Second, we present an end-to-end trainable model that integrates the feature extraction, text encoding, proposal generation, and proposal-text alignment. Our proposed model is evaluated on a simple dataset, demonstrating its ability to improve the semantic alignment between the text and object proposals, leading to more relevant results in semantic image search.

The remainder of this paper is organized as follows: Section II explores the related works. Section III provides a detailed description of the proposed model, including the architecture and the training process. Section IV presents the experimental results, including a comparison with existing object detection models. Finally, Section V concludes the paper and discusses future work.

% Related Work
\section{Related Work}
This sections provides the background and context for the proposed solution by reviewing related work in object detection and semantic image search. It covers a brief overview of traditional object detection methods, as well as image-text alignment methods that are used in semantic image searching algorithms.

\subsection{Object Detection}
The development of object detection algorithms has its roots in the early days of computer vision. The first algorithms focused on detecting simple objects such as edges and corners, such as the edge detection algorithm proposed by Canny \cite{canny1986computational}, which uses the gradient of image intensity to detect edges. Another early object detection algorithm is the Hough Transform \cite{illingworth1988survey}, a classic technique for detecting lines, circles, and other shapes in an image by transforming the image into a parameter space. The Viola-Jones face detection algorithm \cite{viola2001rapid} uses a set of Haar-like features combined in a cascade of classifiers to describe the appearance of a face.

The advancement of object detection algorithms has been significantly influenced by the integration of machine learning techniques. Among these techniques, Support Vector Machines (SVM) \cite{felzenszwalb2009object} and Histogram of Oriented Gradients (HOG) \cite{dalal2005histograms} have emerged as the primary approach for object detection. SVM functions as a linear classifier, training a binary classifier to distinguish between object and non-object regions in an image by maximizing the margin between positive and negative examples in the training set. HOG, on the other hand, acts as a feature descriptor, capturing the shape and texture information of an object by computing histograms of gradient orientation in small cells of an image. These algorithms offer a more automated and scalable way of learning object representations, resulting in the detection of a wider range of object categories with improved accuracy.

In 2012, the Convolutional Neural Networks \cite{krizhevsky2017imagenet} became more feasible to learn more robust and high level features representations. Several object detection models based on CNNs were proposed. These models usually can be categorized in: (i) two-stage detectors, where a set of proposals are first generated and then passed to a separate network to classify and refine the object locations; and (ii) one-stage detectors, which performs object detection in a single stage, increasing its speed while reducing accuracy \cite{zou2023object}.

In 2014, Girshick et al. \cite{girshick2014rich,girshick2015region} proposed a two-stage detector, the Regions with CNN features (RCNNs), in which by using selective search \cite{uijlings2013selective} a set of object proposals is extracted, and then each proposal is rescaled to a fixed size image and sent to a pre-trained CNN in order to have its features extracted. Finally, a SVM classifier is used to detect the presence of an object and predict its class. R-CNN yield a significant improvement in the mean Average Precision with IoU threshold of 0.5 (mAP@.5) of 58.5\% on VOC07 dataset. However, the high number of proposals, around 2000 for a single image, resulted in a extremely slow detection speed.

To address this challenge, another CNN based object detection models were proposed. In 2014, the SPPNet was introduced by He et al. \cite{he2015spatial} and aimed to address the issue of varying object scales in object detection tasks. The SPPNet uses a spatial pyramid pooling layer to aggregate information from different regions of an image, allowing the network to handle objects of different scales effectively, achieving a mAP of 59.2\% on VOC07, more than 20 time faster than R-CNN. On the other hand, Fast R-CNN, introduced by Girshick in 2015 \cite{girshick2015fast}, is a more efficient variant of R-CNN. Fast R-CNN uses a shared convolutional layer for both region proposals and final classification, leading to faster processing times compared to previous methods and achieved a mAP@.5 of 70.0\% on VOC07 with a detection speed over 200 times faster, compared to R-CNN.

Later, an improved version of Fast RCNN, called Faster RCNN was proposed by Ren et al. \cite{ren2015faster} which introduced the Region Proposal Network (RPN) to efficiently generate object proposals and the shared convolutional layer for both proposal generation and object classification, leading to faster processing times and high accuracy in object detection tasks when compared to the original R-CNN. Faster R-CNN achieved a mAP@.5 of 42.7\% on COCO and 73.2\% on VOC07 and 17 fps with ZF-NET.

In 2015, Redmon et al. \cite{redmon2016you} introduced You Only Look Once (YOLO). YOLO was the first one-stage detector in the deep learning era, and it immediately stood out for its incredible speed, with a fast version running at 155 fps and a VOC07 mAP@.5 of 52.7\%. YOLO operates by applying a single neural network to the full image, dividing the image into regions and predicting bounding boxes and probabilities for each region simultaneously. While its high-speed detection was groundbreaking, YOLO suffered from a drop in localization accuracy compared to two-stage detectors, particularly for small objects. Nevertheless, subsequent versions of YOLO, such as YOLOv7 \cite{wang2022yolov7}, have addressed this issue through techniques like dynamic label assignment and optimized structures, achieving outstanding speed and accuracy ranges from 5 to 160 fps.

Single-Shot Multibox Detector (SSD), proposed by Liu et al. \cite{liu2016ssd} in 2015, brought a significant improvement in detection accuracy to one-stage detectors. SSD introduced multireference and multiresolution detection techniques, which notably improved the accuracy of one-stage detectors, particularly for small objects. SSD offered advantages in both detection speed, achieving 59 fps, and mAP@.5 of 46.5\% on COCO. The key difference between SSD and previous detectors is that SSD performs detection on different layers of the network for objects of different scales, whereas previous detectors only operated on their top layers.

RetinaNet, introduced by Lin et al. \cite{lin2017focal} in 2017, tackles the accuracy issues that one-stage detectors had faced compared to two-stage detectors. The authors identified a significant foreground-background class imbalance during the training of dense detectors as the main cause of this gap. To address this, RetinaNet introduced a new loss function named focal loss, which reshaped the standard cross entropy loss to put more focus on hard, misclassified examples during training. With focal loss, RetinaNet was able to achieve a comparable mAP@.5 of 59.1\% on COCO, with two-stage detectors while maintaining a high detection speed.

\subsection{Image-Text Alignment}
Image-text alignment is a field of study that explores the relationship between images and their textual description. This field has gained significant attention in recent years due to its wide range of applications, such as visual captioning, visual question answering, and cross-modal retrieval.

In 2019, Gu et al. proposed ImageBERT \cite{qi2020imagebert}, a pre-trained deep learning model that learns to understand the relationship between images and text by leveraging a large-scale image-text corpus. ImageBERT is based on the BERT (Bidirectional Encoder Representations from Transformers) \cite{devlin2018bert} architecture and has been pre-trained on large-scale visual and textual data. The model uses a combination of self-attention mechanisms and fully-connected layers to learn to align visual and textual information in a common semantic space. The pre-training step allows ImageBERT to achieve high performance on a variety of vision and language tasks without the need for task-specific fine-tuning.

CLIP (Contrastive Language-Image Pretraining), introduced by Radford et al. \cite{radford2021learning} in 2021, is another multi-modal architecture designed to learn a common semantic representation between text and images. It uses a large pre-training corpus of textual and visual data and a contrastive learning objective to learn how to align text and image data. The model has two encoders, one for text and one for images, that operate in parallel and are optimized together. The two encoders are trained to predict whether a text description is likely to match the image it is paired with. CLIP has been shown to perform well on various tasks such as cross-modal retrieval, text-to-image generation, and visual question answering.

Also in 2021, ALIGN (Attentive Language-Image Generative model) is a deep learning model proposed by C. Jia et al. \cite{jia2021scaling}. ALIGN is designed to align text and image modalities through an attention mechanism. The model has two encoders for text and images that operate in parallel, and a decoder that combines their representations using an attention mechanism to generate an aligned representation. The attention mechanism allows the model to focus on the most important regions of the image and text representations and to align the two modalities in a meaningful way. The model is trained to reconstruct the input image and text in an autoencoder-style architecture, which allows it to learn to preserve the information that is important for both modalities.

% Methodology
\section{Methodology}
\begin{figure}[h!]
  \centering
  \includegraphics[width=1.0\linewidth]{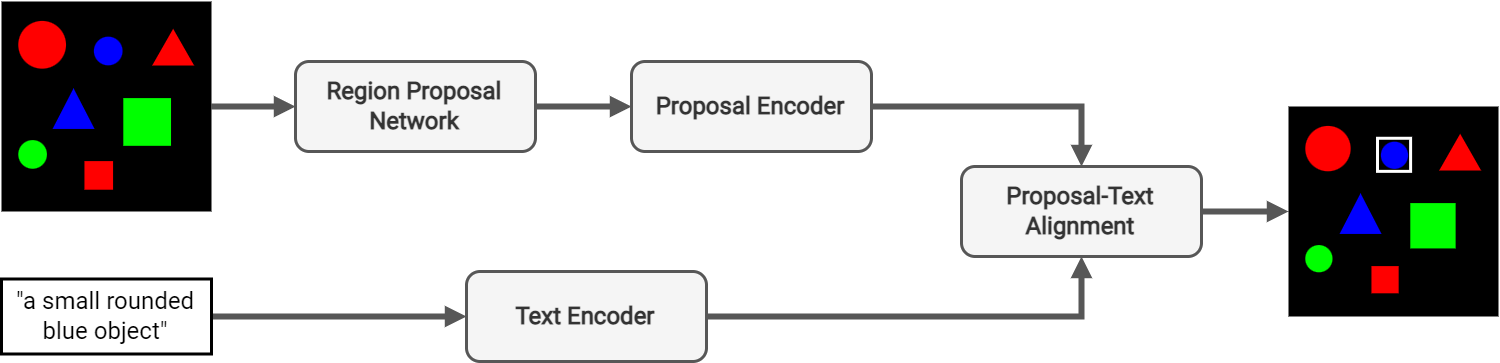}
  \caption{Architecture of the proposed model.}
  \label{fig:architecture}
\end{figure}

This section presents the details of our proposed approach for semantic image search. It covers the components of our model, including the Region Proposal Network (RPN), the proposal and text encoders, and the proposal-text alignment model, as shown in Figure \ref{fig:architecture}. We also explain our approach for alignment using contrastive learning, which is a crucial aspect of our method. The section concludes with a description of the end-to-end training process and the experimental setup.

\subsection{Region Proposal Network}
Region Proposal Networks (RPNs) \cite{ren2015faster} play a significant role in object detection systems. Their purpose is to produce region proposals that have the potential to contain object instances within an image. These proposals are then handed off to a later stage, like a Fast R-CNN network, for further classification and refinement.

\begin{figure}
  \centering
  \includegraphics[width=1.0\linewidth]{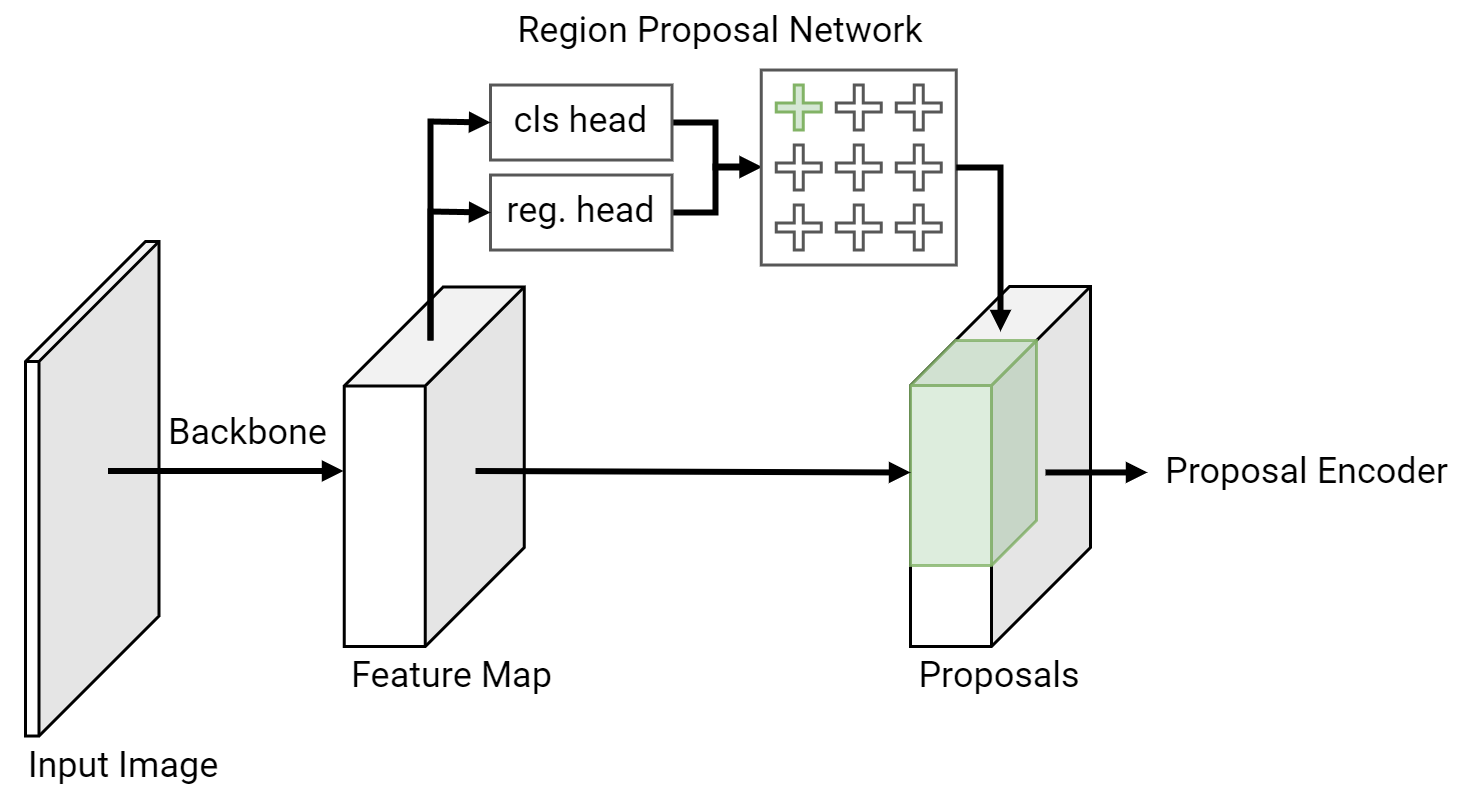}
  \caption{Inner architecture of the Region Proposal Network module.}
  \label{fig:rpn}
\end{figure}

The RPN functions on the feature maps that result from the convolutional layers in a deep neural network. Given an input image, the RPN generates multiple region proposals, each defined by a set of anchor boxes. Anchors are fixed-size boxes with specified aspect ratios that are densely placed over the feature maps. The RPN predicts a binary classification label for each anchor box to determine whether it contains an object or not, and also predicts bounding box regression coefficients to refine the anchor boxes into the final proposals.

The process for training the binary classification head of RPNs involves assigning binary class labels, indicating if it is an object or non-object, to each anchor. Positive labels are given to anchors that meet one of two criteria: (i) having the highest Intersection-over-Union (IoU) overlap with a ground-truth box, or (ii) having an IoU overlap higher than a maximum threshold, such as 0.7, with any ground-truth box. Although the second condition is typically enough to identify positive samples, the first condition is also implemented for cases when no positive samples are found through the second criterion. Negative labels are assigned to anchors that have an IoU ratio lower than a minimum threshold, such as 0.3, with all ground-truth boxes. Anchors that do not meet either positive or negative criteria do not play a role in the training. However, it's worth noting that the number of negative labels usually surpasses the number of positive samples, potentially leading to a bias towards negative samples. To counteract this, negative samples are randomly selected so that the ratio of positive to negative samples is roughly 1:1.

For training the regression head of RPNs to predict the bounding boxes, only positive samples are used and the outputs are parameterized by:

\begin{align}
\begin{split}
t_{\mathrm{x}} & =\left(x-x_{\mathrm{a}}\right) / w_{\mathrm{a}}, \quad t_{\mathrm{y}}=\left(y-y_{\mathrm{a}}\right) / h_{\mathrm{a}} \\
t_{\mathrm{w}} & =\log \left(w / w_{\mathrm{a}}\right), \quad t_{\mathrm{h}}=\log \left(h / h_{\mathrm{a}}\right) \\
t_{\mathrm{x}}^* & =\left(x^*-x_{\mathrm{a}}\right) / w_{\mathrm{a}}, \quad t_{\mathrm{y}}^*=\left(y^*-y_{\mathrm{a}}\right) / h_{\mathrm{a}}, \\
t_{\mathrm{w}}^* & =\log \left(w^* / w_{\mathrm{a}}\right), \quad t_{\mathrm{h}}^*=\log \left(h^* / h_{\mathrm{a}}\right)
\end{split}
\end{align}

where $x$, $y$, $w$, and $h$ represent the center coordinates and the width and height of the box, respectively. $x_{\mathrm{a}}$, $y_{\mathrm{a}}$, $w_{\mathrm{a}}$, and $h_{\mathrm{a}}$ denote the corresponding values of the anchor box, and $x^*$, $y^*$, $w^*$, and $h^*$ for the ground-truth box.

Therefore, the objective of the RPN is to minimize a multi-task loss function that encompasses a component for the binary classification task and another component for the regression task:

\begin{equation}
\mathcal{L}_{\text{rpn}} = \frac{1}{N_{cls}} \sum_{i} L_{cls}(p_i, p_{i}^{*}) + \lambda \frac{1}{N_{reg}} \sum_{i} p_i^{*} L_{reg}(t_i, t_i^{*})
\end{equation}

where $N_{cls}$ and $N_{reg}$ are the total number of anchors used for classification and regression, respectively, $p_i$ and $t_i$ are the predicted class label and bounding box regression coefficients for the $i^{th}$ anchor, and $p_{i}^{}$ and $t_{i}^{}$ are the ground truth labels. $L_{cls}$ represents the binary cross-entropy loss for classification, and $L_{reg}$ represents the smooth L1 loss for bounding box regression. The hyperparameter $\lambda$ balances the two components.

The object proposals generated by the RPN are then filtered and refined using non-maximum suppression (NMS) and bounding box regression. NMS eliminates overlapping proposals, while bounding box regression adjusts the dimensions and position of the remaining proposals. The result of this process is a set of high-quality object proposals, which will be fed into the proposal encoder.

\subsection{Proposal Encoder}
The Proposal Encoder is used to map each region proposal, $r_j$, predicted by the RPN, into a fixed sized embedding vector $f_r(r_j)$ from an embedding vector space $\mathbb{R}^{d_r}$. We aim for the embedding vectors produced by the Proposal Encoder to capture both spatial and semantic information within each proposal, allowing the proposal-text alignment model to make informed decisions about the relationships between objects and their textual descriptions.

\begin{figure}
  \centering
  \includegraphics[width=0.75\linewidth]{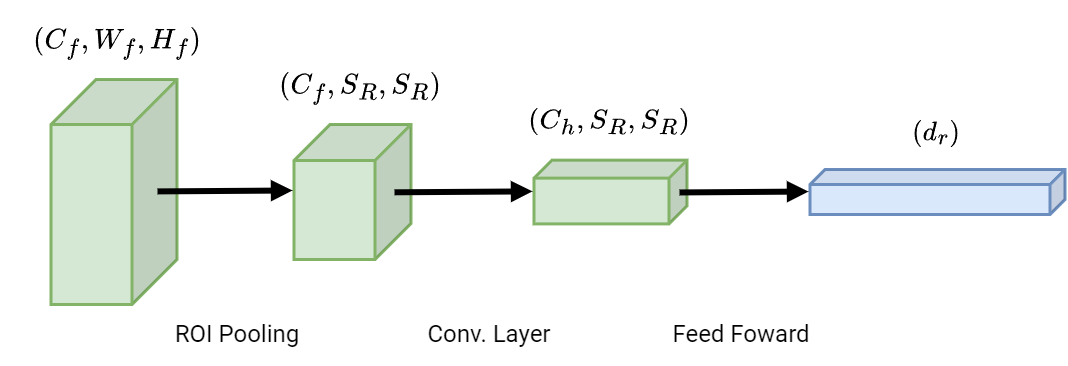}
  \caption{Inner architecture of the Proposal Encoder module.}
  \label{fig:proposal_encoder}
\end{figure}

The architecture of the Proposal Encoder Module is comprised of three main components: ROI pooling, a convolutional layer, and two fully connected layers, as shown in Figure \ref{fig:proposal_encoder}. The ROI pooling layer resizes each region proposal with shape $(C_f,W_f,W_f)$ to a fixed size, $(C_f,S_R,S_R)$, while preserving spatial information. This output is then fed into a convolutional layer with output channels dimension of $C_h$, which further extracts and refines the features of the proposals. The final output of the module is the embedding vectors, produced by the two fully connected layers. Additionally, the embedding vectors are constrained to live in a $d_r$-dimensional hyper-sphere through L2-normalization.

The size of the encoded proposals, $d_r$, is a hyperparameter that can be adjusted depending on the needs of the object detection task. In this way, the Proposal Encoder module effectively converts region proposals into compact and informative feature representations.

\subsection{Text Encoder}
\begin{figure}[h!]
  \centering
  \includegraphics[width=0.70\linewidth]{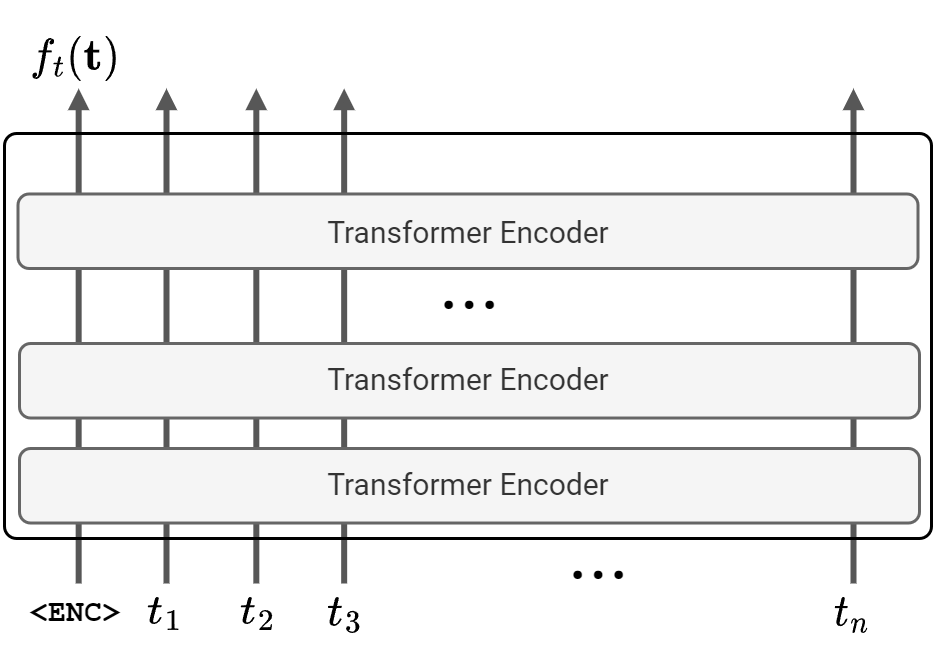}
  \caption{Inner architecture of the Text Encoder module using a stack of Transformer encoders.}
  \label{fig:text_encoder}
\end{figure}

The Text Encoder is designed to transform the textual descriptions of the objects of interest in the image into a fixed-sized dense vector representation, $f_t(\textbf{t}) \in \mathbb{R}^{d_t}$, which will later be utilized in the Alignment Model. Given a sequence of tokens $\textbf{t}=(t_0,t_1,...,t_n)$ represented as word embeddings, the Text Encoder carries out the encoding mechanism through the use of a Transformer encoder architecture, as shown in Figure \ref{fig:text_encoder}. The goal is to extract contextual information from the input text sequence.

A special token, denoted as \texttt{<ENC>}, is added into the start of the input sequence to serve as an aggregated representation of the entire input text, similar to the [CLS] token in BERT. This allows the Text Encoder to effectively condense the textual information into a compact, fixed-sized vector representation.

The size of the encoded textual representation, $d_t$, is a hyperparameter that can be adjusted depending on the needs of the object detection task. The Text Encoder module effectively converts textual descriptions into compact and informative feature representations that are suitable for use in the Alignment Model. Additionally, the embedding vectors are constrained to live in a $d_t$-dimensional hyper-sphere through L2-normalization.
% The Text Encoder is designed to transform the textual descriptions of the objects of interest in the image into a fixed-sized dense vector representation, as shown in Figure \ref{fig:text_encoder}. For this, given a sequence of tokens $\textbf{t}=(t_0,t_1,...,t_n)$ represented as word embeddings, the Text Encoder carries out the encoding mechanism through the use of a Transformer Encoder architecture to extract the contextual information from the text sequence into the embedding vector $f_t(\textbf{t}) \in \mathbb{R}^{d_t}$. Futhermore, a special token, denoted as \texttt{<ENC>}, is added into the start of the input sequence to serve as an aggregated output and the output is constrained by a L2-normalization.

% [Image here]

% The size of the encoded text, $d_t$, is also a hyperparameter that can be adjusted depending on the needs of the object detection 

% The output of the Text Encoder Network, $f(t) \in \mathbb{R}^{d_t}$, represents the final embedding of the text sequence, which is used in conjunction with the vector embeddings of the proposals obtained from the Proposal Encoder Network in the proposal-text alignment model.

\subsection{Proposal-Text Alignment}
\begin{figure}[h!]
  \centering
  \includegraphics[width=0.8\linewidth]{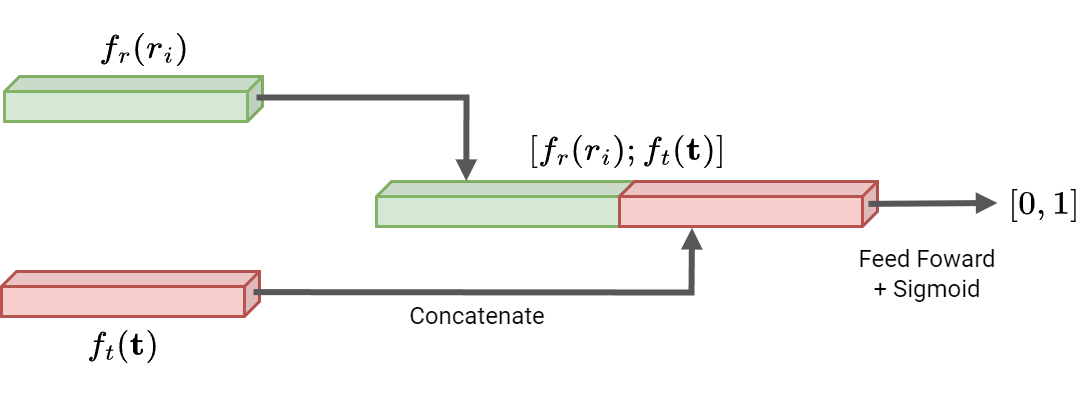}
  \caption{Inner architecture of the Proposal-Text Alignment module.}
  \label{fig:proposal_text_alignment}
\end{figure}

The Alignment Model is the final component of the proposal-text alignment framework, which aligns the proposals' embeddings obtained from the Proposal Encoder with the textual embeddings obtained from the Text Encoder. The aim of this model is to predict the similarity between each proposal and the textual description, which can then be used to select the most relevant proposals for a given text description. The inputs of the model are the embedding vectors for a given proposal and a given text, and the output is a value between zero and one, indicating the match between them, as shown in Figure \ref{fig:proposal_text_alignment}.

Given the proposal embedding $f_r(r_j) \in \mathbb{R}^{d_r}$ and the text embedding $f_t(\textbf{t}) \in \mathbb{R}^{d_t}$, the concatenated representation $z_{j} = [f_r(r_j); f_t(\textbf{t})] \in \mathbb{R}^{d_r + d_t}$ is passed to a two-layers fully connected neural network, with the first layer having output dimension of $d_j$ with a ReLU activation and the second layer with output dimension 1 with a sigmoid activation function, representing an alignment score between zero and one, as:

\begin{equation}
    \textrm{Align}(r_j,\textbf{t})=\sigma(\max(0, z_jW_1+b_1)W_2+b_2)
\end{equation}

where $W_1 \in \mathbb{R}^{d_r+d_t,d_j}$, $b_1 \in \mathbb{R}^{d_j}$, $W_2 \in \mathbb{R}^{d_j,1}$ and $b_2 \in \mathbb{R}^{1}$ are the parameters of the neural network and $\sigma(\cdot)$ is the sigmoid function.

The Alignment Model can be trained end-to-end using a suitable loss function, such as cross-entropy loss, to maximize the similarity between the positive proposal-text pairs and minimize the similarity between the negative proposal-text pairs. The output of the Alignment Model can be used for various tasks, such as visual grounding, where the goal is to locate a specific object in an image given a textual description.

The loss function for the Alignment Model is designed to measure the difference between the predicted relationship between a proposal and a textual description, and the ground truth relationship. The prediction is made through the Alignment Model's binary classification output, which represents the confidence of a given proposal being described by the textual description. The loss is calculated by taking the binary cross-entropy between the predicted probability of the positive class and the ground truth label, which is either 1 (positive) or 0 (negative). The overall loss function is the sum of the binary cross-entropy losses for all the positive and negative proposal-text pairs. The goal is to minimize this loss, which will result in the Alignment Model becoming more accurate in predicting the relationship between proposals and their textual descriptions. By minimizing the loss, the Alignment Model learns to discriminate between proposals that are and are not described by a textual description, leading to improved object detection performance.

\begin{equation}
\mathcal{L}_{\text{align}} = -\frac{1}{N} \sum_{i=1}^N (y_i \log(\hat{y_i}) + (1-y_i)\log(1-\hat{y_i}))
\end{equation}

where $N$ is the number of proposals, $\hat{y_i}$ is the predicted alignment score for proposal $i$, and $y_i$ is the true label indicating if the proposal $i$ is aligned with the corresponding textual description or not. The loss function optimizes the model to predict binary alignment scores that are as close as possible to the ground-truth labels.

\subsection{Scoring Proposals}
The scoring proposals stage is the final step in the proposal-text alignment model. Given a set of region proposals, $\textbf{r}$, and a textual description $\textbf{t}$, the scoring proposals stage computes a score, $\textrm{Score}(r_j,\textbf{t})$, for each proposal $r_j$ that indicates the level of semantic compatibility between the proposal and the text description.

The score calculation is based on the outputs from the Alignment Model, $\textrm{Align}(r_j,\textbf{t})$. These outputs are interpreted as the probabilities of each proposal $r_j$ being the correct match for the textual description $\textbf{t}$. The scores for all the proposals are then combined and ranked, with the proposals with the highest scores being selected as the final detections. The equation for the Score function can be written as follows:

\begin{equation}
\textrm{Score}(r_j,\textbf{t})= \textrm{Conf}(r_j) \times \textrm{Align}(r_j,\textbf{t})
\end{equation}

where $\textrm{Conf}(r_j)$ is the confidence score of the region proposal $r_j$ obtained from the RPN, and $\textrm{Align}(r_j,\textbf{t})$ is the alignment score obtained from the Alignment Model. The final detections are selected by selecting the top-k proposals with the highest scores, where $k$ is a hyperparameter that can be adjusted depending on the desired level of precision and recall.

\subsection{End-to-End Training}
The end-to-end training of the model involves training both the RPN and the Alignment Model jointly. Each training example consists of a single image, a set of bounding boxes for the objects in the image; a text query and an indicator of the alignment of the text for each object, indicating which objects are positives or negatives depending on the text query.

To train the RPN we follow the process described in \cite{ren2015faster}, where the classification and regression heads are trained to predict the confidence and the regression boxes of each object, respectively. The bounding boxes of the positive samples from the RPN training step are used to extract the parts of interests from the feature map and passed to the Proposal Encoder, while the text query from the given training example is encoded through the Text Encoder. Subsequently, the proposals and text encodings are passed on to the Alignment Model, which calculates scores that are used in the computation of the alignment loss. The loss is then backpropagated to both the inputs of the model, updating the weights of the model during the end-to-end training process.

\newpage
% Experiments and Resutls
\section{Experiments and Results}
This section presents the experimental setup and results obtained in the evaluation of the proposed method. The experiment was conducted using simple images, with a single training session. The results show that the proposed method is effective in performing semantic image search and generating semantically relevant object proposals, offering promising performance compared to traditional object detection models. In the following sections, the generation of the training data, the experimental setup, the evaluation metrics, and the results are discussed in detail, providing insights into the effectiveness of the proposed method in solving the task of semantic image search within images.

\subsection{Data}
\begin{figure}[h!]
  \centering
  \includegraphics[width=1.0\linewidth]{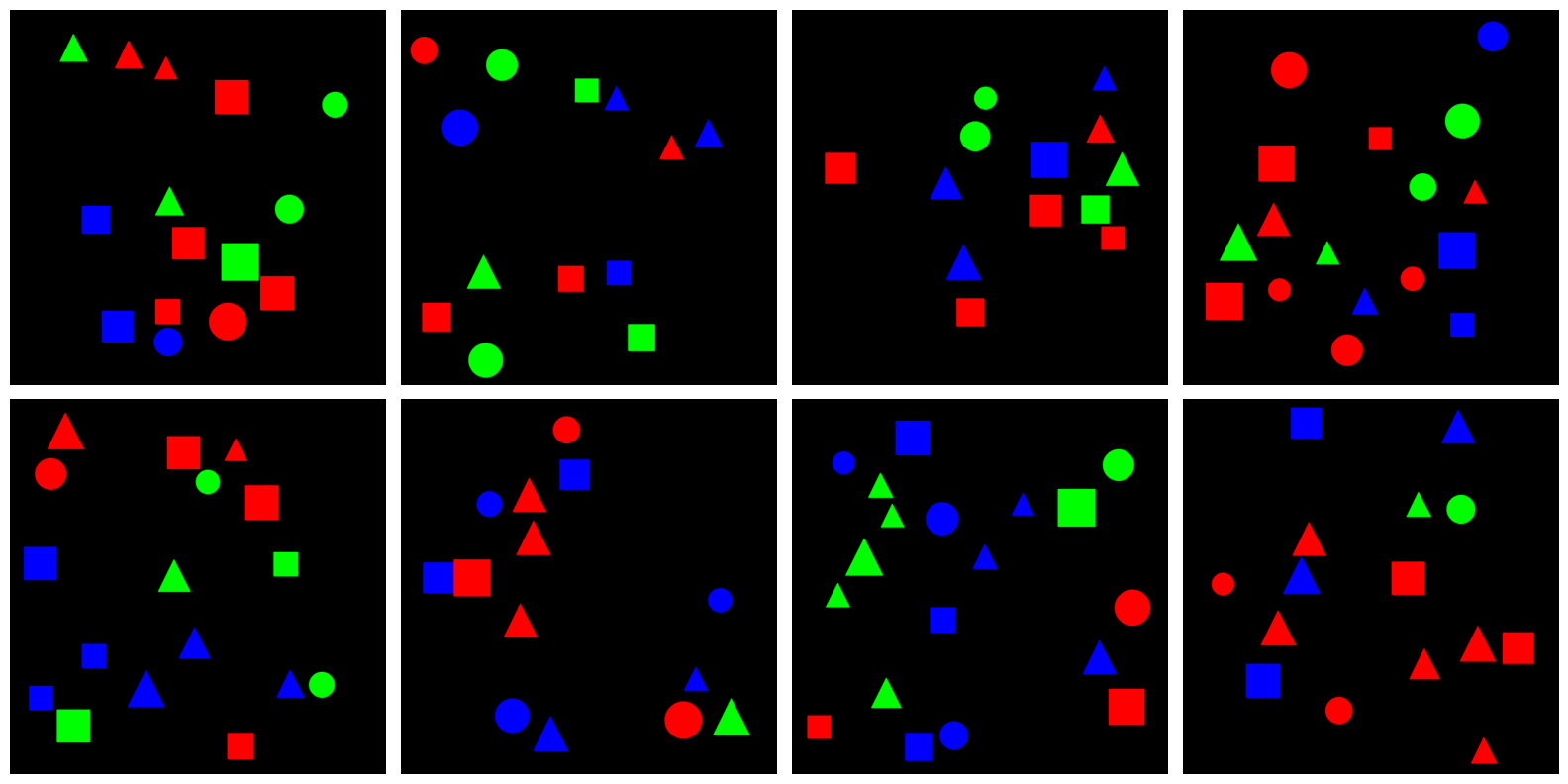}
  \caption{Examples of images used in training.}
  \label{fig:data}
\end{figure}

The data used for training in this implementation consisted of 5000 examples generated synthetically, where 4000 were used for training, 500 for validation and 500 for testing. Each example consists of a 512x512 RGB image, containing 10 to 20 randomly placed shapes, including circles, squares, and triangles, with randomly assigned colors including red, green, and blue on a black background. The objects in the images were annotated with bounding boxes, and a sentence in natural language was generated as the text query to describe the objects. Additionally, an annotation indicating which objects were aligned with the text was provided to serve as the ground truth. Figure \ref{fig:data} shows a few examples of the images used in training.

\subsection{Implementation details}
For this implementation, we used a pre-trained MobileNetV2 \cite{sandler2018mobilenetv2} with its classification head removed, remaining only the CNNs with an output channels size of 1280. In the Region Proposal Network we defined a set of equally spaced anchors, alined with the feature map, with equal widths and heights of 64px. The maximum and minimum IoU thresholds which will defined the positive and negative samples are 0.6 and 0.1 respectively. The Proposal Encoder have the ROI pooling output size of (2,2), and the embedding dimension for the outputs is 64. The Text Encoder consists of a Transformer encoder with embedding dimension of 64, 2 heads and 1 layer. The Proposal-Text Alignment model have a hidden dimension of 64.

\subsection{Training}
\begin{figure}[h!]
  \centering
  \includegraphics[width=1.0\linewidth]{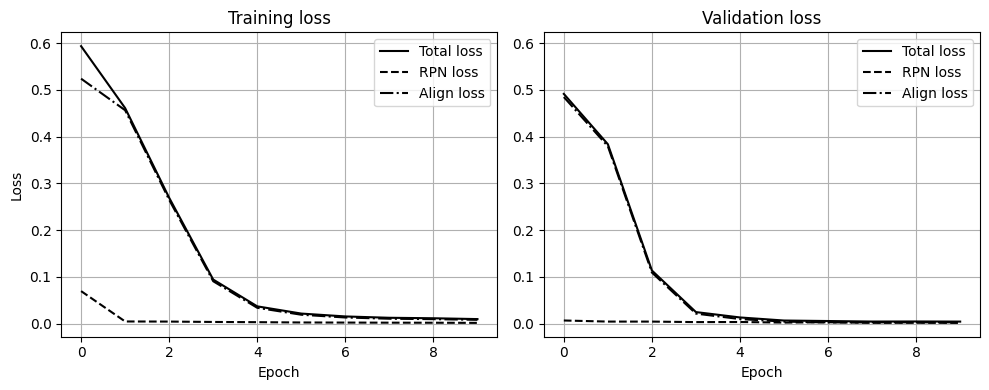}
  \caption{Losses curves for training and validation.}
  \label{fig:loss}
\end{figure}

The training process was carried out over 10 epochs, with a batch size of 32. The optimization algorithm used was Stochastic Gradient Descent (SGD) with a learning rate of 1e-2, momentum of 0.9, and weight decay of 1e-5. In addition, a step-based learning rate scheduler with step size 3 and gamma 0.9 was employed to adjust the learning rate dynamically during the training process. This helped to mitigate the risk of overfitting and converge to a better optimal solution. The curves for the RPN loss and Align loss can be seen in Figure \ref{fig:loss}. It is observed that the RPN loss occurs rapidly converges, while the total loss is primarily impacted by the Align loss. This phenomenon can be attributed to the nature of the Align loss, which measures the semantic alignment between the object proposals and the text query, and hence has a more complex optimization problem than the RPN loss. The interplay between these two losses ultimately determines the overall performance of the proposed model.

\subsection{Quantitative Results}
The experiments were divided into two groups. The first group comprised the evaluation of the results obtained from the ground truth and proposals bounding boxes of the RPN, i.e. all the predicted boxes, not considering the text query. The second group consisted of the evaluation of the results obtained from the filtered proposals and bounding boxes that were aligned with text. A threshold of 0.5 was defined to determine true and false positives and negatives. The metrics evaluated were the precision, recall, IoU and align accuracy. The tests were conducted on a test set of 500 examples. Table \ref{tab:results} shows the obtained results.

\begin{table}[h!]
    \caption{Experiment Results}
    \centering
    \begin{tabular}{lcc}
    \toprule
    \multicolumn{1}{c}{Metric} & \multicolumn{1}{c}{All Proposals} & \multicolumn{1}{c}{Aligned Proposals} \\
    \midrule
    Mean Precision & 0.932 & 0.948 \\
    Mean Recall & 0.862 & 0.966 \\
    Mean IoU & 0.759 & 0.767 \\
    Alignment Accuracy & N/A & 0.996 \\
    \bottomrule
    \end{tabular}
    \label{tab:results}
\end{table}

For the mean precision metric, it was calculated as the ratio of true positive bounding boxes to the total number of positive bounding boxes predicted by the model. This metric reflects the ability of the model to accurately predict the positive bounding boxes in the image, taking into consideration both true positive and false positive predictions. The higher the mean precision, the more accurate the models's predictions of positive bounding boxes are, which leads to better object detection performance. The results were 0.932 for all the proposals predicted by the models and 0.948 for the proposals aligned with the text query. It can be observed that the mean precision increased in the second set of experiments, this is because that by removing proposals that were not aligned with the text, the total number of positive predictions is reduced, thus reducing the chance of false positive predictions.

For the mean recall metric, it was calculated as the ratio of true positive bounding boxes to the total number of ground truth bounding boxes. This metric reflects the ability of the model to detect all the positive instances in the image, taking into consideration both true positive and false negative predictions. The higher the mean recall, the better the model is at detecting all positive instances in the image, resulting in improved object detection performance. The results were 0.862 for all the proposals predicted by the models and 0.966 for the proposals aligned with the text query. It can be observed that the mean recall increased in the second set of experiments, this is because by filtering out proposals that were not aligned with the text, the model has a better chance of detecting all positive instances, thus increasing the mean recall.

\begin{figure}[h!]
  \centering
  \includegraphics[width=0.7\linewidth]{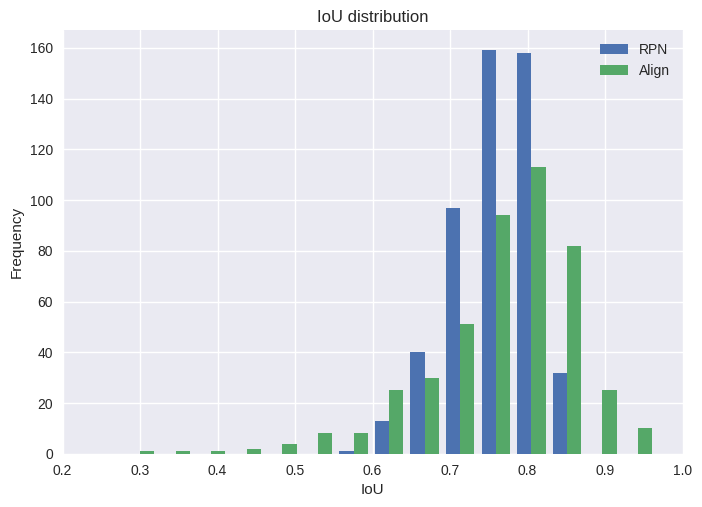}
  \caption{Histograms for the IoU for all proposals and filtered proposals based on text alignment.}
  \label{fig:iou_hist}
\end{figure}

The Intersection over Union (IoU) metric measures the overlap between the ground truth bounding box and the predicted bounding box. The IoU is calculated as the ratio of the area of overlap between the two boxes and the area of their union. A high IoU value indicates a high level of overlap and hence, a better prediction by the model. In the experiments, the mean IoU was calculated by taking the average of the IoUs of all the bounding boxes predicted by the RPN. The results were 0.759 for all the proposals predicted by the RPN and 0.767 for the proposals aligned with the text query. The results show that the mean IoU values for both sets of experiments were close, which indicates that filtering the proposals based on alignment with text did not have a significant impact on the mean IoU values. The histograms of the IoUs for both experiments sets are shown in Figure \ref{fig:iou_hist} also supports this observation, with the majority of the IoUs falling in the range of 0.7 to 0.8.

The results obtained indicate that the proposed method performs well in terms of precision, recall and IoU, with a higher mean precision and recall in the second group. The high align accuracy further confirms the effectiveness of the proposed alignment mechanism. These results demonstrate the potential of the proposed method for semantic image search tasks.

\subsection{Qualitative Results}
\begin{figure}
  \centering
  \includegraphics[width=0.8\linewidth]{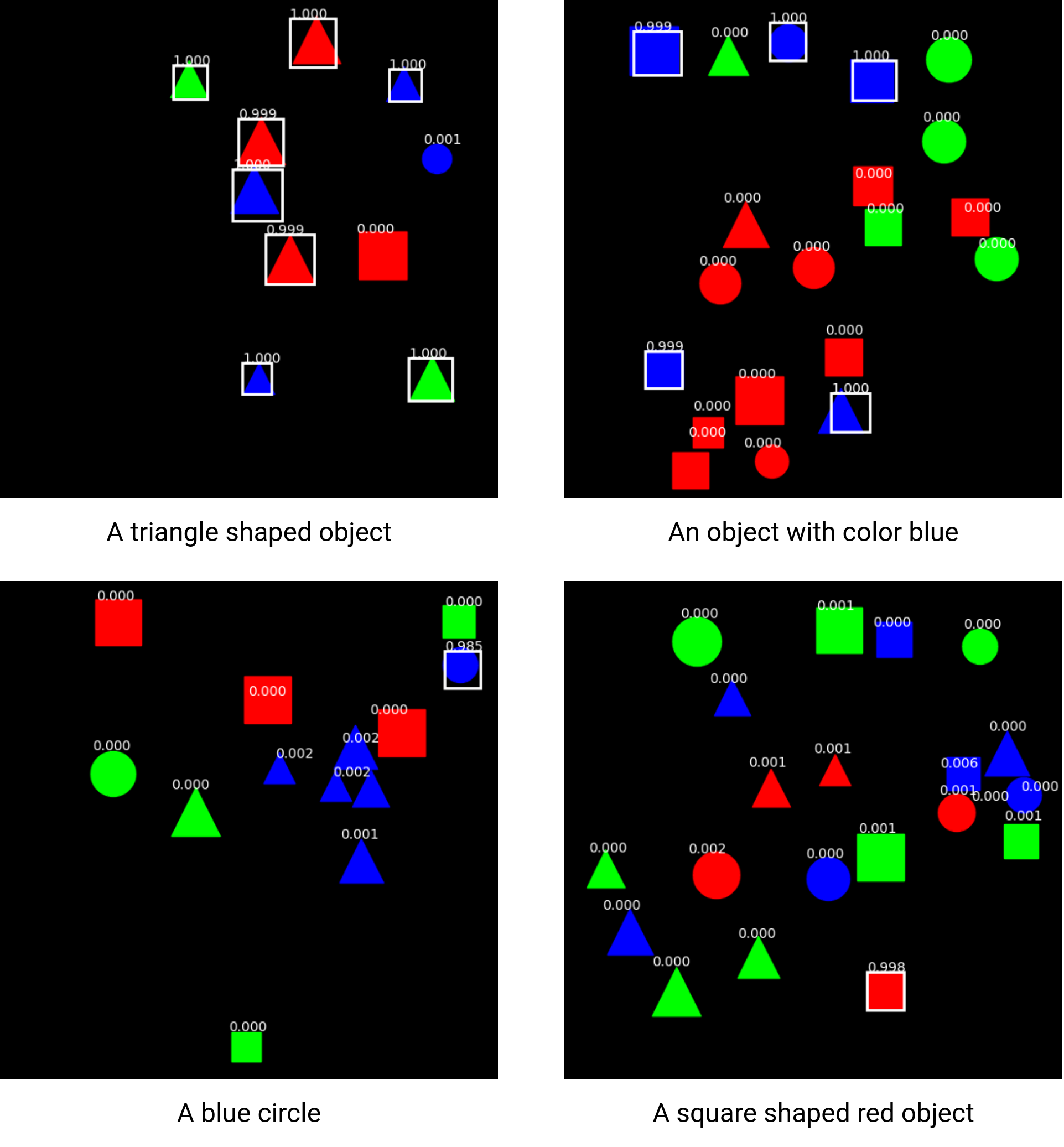}
  \caption{Histograms for the IoU for all proposals and filtered proposals based on text alignment.}
  \label{fig:output_examples}
\end{figure}

The qualitative results of the experiments revealed the effectiveness of both the RPN and the text-filtered proposals in providing accurate bounding boxes on the objects that matched the text query. To demonstrate the performance of the models, multiple examples were displayed, with each example comprising of an image and its corresponding text query. The images in the examples contained bounding boxes that were superimposed on the objects that matched the text query. These bounding boxes were generated by the RPN and filtered accordingly to the Proposal-Text alignment. Several examples were presented to illustrate the performance of the models, with each image accompanied by a corresponding text query. The bounding boxes were drawn on the image only if the score was higher than 0.5, ensuring that the predictions were confident enough to be considered true positive detections. The results are shown in Figure \ref{fig:output_examples}.

\newpage
\section{Conclusion}
In conclusion, this paper presented a novel approach to semantic image search, which is an important task in computer vision. The proposed model leverages a single feature extractor, a pre-trained Convolutional Neural Network, and a transformer encoder to encode the text query and perform proposal-text alignment using contrastive learning. The end-to-end training process of the model allowed for an efficient and effective solution for semantic image search. The qualitative results were consistent with the quantitative results and demonstrated the effectiveness of the text-filtered proposals in providing accurate bounding boxes on the objects that matched the text query. The results showed that the proposed method produced promising results when trained with simple images, demonstrating its potential for further improvement and refinement. Overall, the proposed method provides a step forward in the field of semantic image search and highlights the importance of considering semantic alignment in object detection models.

In future works, we aim to evaluate the proposed method on a larger and more diverse dataset to further test its performance and robustness. Additionally, we plan to explore the use of pre-trained text encoders such as BERT, which has shown to be effective in various NLP tasks. The integration of these pre-trained text encoders may further improve the semantic alignment between the image and text, leading to better object detection performance. Furthermore, we plan to investigate different methods for proposal-text alignment, including pre-trained models such as CLIP, to further optimize the performance of the model. By exploring these avenues, we hope to provide a more robust solution for semantic image search and advance the field of computer vision.

\newpage

% \section*{Acknowledgments}
% This was was supported in part by......

%Bibliography
\bibliographystyle{unsrt}  
\bibliography{references}

\end{document}